%% file: main.tex
\documentclass[runningheads]{llncs}
\usepackage{graphicx}
\usepackage{tikz}
\usepackage{comment}
\usepackage{amsmath,amssymb} % define this before the line numbering.
\usepackage{color}
\usepackage{tabularx}
\usepackage{booktabs}
\usepackage{colortbl}
\usepackage{tabu}
\usepackage{multirow}
\usepackage{setspace}
\usepackage[skip=1ex,font=small,labelsep=period]{caption}

\usepackage[pagebackref=true,breaklinks=true,letterpaper=true,colorlinks=true,urlcolor=bblue,bookmarks=false,allcolors=black]{hyperref}

\makeatletter
\renewcommand{\fnum@figure}{Figure \thefigure}
\makeatother

\newcolumntype{Y}{>{\centering\arraybackslash}X}
\newcolumntype{R}{>{\raggedleft\arraybackslash}X}
\newcolumntype{L}{>{\raggedright\arraybackslash}X}

\newcommand{\mytilde}{\raise.17ex\hbox{$\scriptstyle\sim$}}

\definecolor{bblue}{rgb}{0.0,0.25,0.65}
\definecolor{ccol}{rgb}{0.9,0.9,0.9}
\definecolor{avgcol}{rgb}{1.0,0.839,0.4}
\definecolor{arcol}{rgb}{0.341,0.733,0.541}
\definecolor{timecol}{rgb}{0.902,0.486,0.451}

\usepackage{xspace}
\newcommand*{\eg}{\emph{e.g.}\@\xspace}
\newcommand*{\ie}{\emph{i.e.}\@\xspace}
\newcommand*{\wrt}{\emph{w.r.t.}\@\xspace}
\newcommand*{\vs}{\emph{vs.}\@\xspace}
\newcommand*{\etal}{\emph{et al.}\@\xspace}

\newcommand\customparagraph[1]{\vspace{0.7em}\noindent\textbf{#1}}

\begin{document}

\pagestyle{headings}
\mainmatter

\title{BOP Challenge 2020 on 6D Object Localization}

\titlerunning{BOP Challenge 2020 on 6D Object Localization}

\author{
 Tom{\'a}{\v{s}}~Hoda{\v{n}}$^{1}$,
 Martin Sundermeyer$^{2}$,
 Bertram Drost$^{3}$,
 Yann Labb{\'e}$^{4}$,\\
 Eric Brachmann$^{5}$,
 Frank Michel$^{6}$,
 Carsten Rother$^{5}$,
 Ji{\v{r}}{\'i}~Matas$^{1}$
}
\institute{
 {$^{1}$Czech Technical University in Prague},
 {$^{2}$German Aerospace Center},
 {$^{3}$MVTec}, \\
 {$^{4}$INRIA Paris},
 {$^{5}$Heidelberg University},
 {$^{6}$Technical University Dresden}
}

\authorrunning{Hoda{\v{n}}, Sundermeyer, Drost, Labb{\'e}, Brachmann, Michel, Rother, Matas}

\maketitle

\begin{abstract}
This paper presents the evaluation methodology, datasets, and results of the BOP Challenge 2020, the third in a series of public competitions organized with the goal to capture the status quo in the field of 6D object pose estimation from an RGB-D image.
In 2020, to reduce the domain gap between synthetic training and real test RGB images, the participants were provided 350K photorealistic training images generated by BlenderProc4BOP, a~new open-source and light-weight physically-based renderer (PBR) and procedural data generator.
Methods based on deep neural networks have finally caught up with methods based on point pair features, which were dominating previous editions of the challenge. Although the top-performing methods rely on RGB-D image channels, strong results were achieved when only RGB channels were used at both training and test time -- out of the 26 evaluated methods, the third method was trained on RGB channels of PBR and real images, while the fifth on RGB channels of PBR images only. Strong data augmentation was identified as a key component of the top-performing CosyPose method, and the photorealism of PBR images was demonstrated effective despite the augmentation.
The online evaluation system stays open and is available on the project website:
\texttt{\href{http://bop.felk.cvut.cz/}{bop.felk.cvut.cz}}.
\end{abstract}

\section{Introduction}

Estimating the 6D pose, \ie the 3D translation and 3D rotation, of rigid objects from a single input image is a crucial task for numerous application fields such as robotic manipulation, augmented reality, or autonomous driving.
The BOP\footnote{BOP stands for Benchmark for 6D Object Pose Estimation~\cite{hodan2018bop}.} 
Challenge 2020 is the third in a series of public challenges that are part of the BOP
project aiming to continuously report the state of the art in 6D object pose estimation.
The first challenge was organized in 2017~\cite{hodan2017sixd} and the results were published in~\cite{hodan2018bop}.
The second challenge from 2019~\cite{hodan2019bop} and the third from 2020 share the same evaluation methodology and leaderboard and the results from both are included in this paper.

Participating methods are evaluated on the 6D object localization task~\cite{hodan2016evaluation}, where the methods report their predictions on the basis of two sources of information. Firstly, at training time, a method is given 3D object models and training images showing the objects in known 6D poses. Secondly, at test time, the
method is provided with a test image and a list of object instances visible in the image, and the goal of the method is to estimate 6D poses of the listed object instances. The training and test images consist of RGB-D (aligned color and depth) channels and the intrinsic camera parameters are known.

The challenge primarily focuses on the practical scenario where no real images are available at training time, only the 3D object models and images synthesized using the models. While capturing real images of objects under various conditions and annotating the images with 6D object poses requires a significant human effort~\cite{hodan2017tless}, the 3D models are either available before the physical objects, which is often the case for manufactured objects, or can be reconstructed at an admissible cost.
Approaches for reconstructing 3D models of opaque, matte and moderately specular objects are well established~\cite{newcombe2011kinectfusion} and promising approaches for transparent and highly specular objects are emerging~\cite{qian20163d,wu2018full,godard2015multi}.

In the BOP Challenge 2019, methods using the depth image channel, which were mostly based on the point pair features (PPF's)~\cite{drost2010model}, clearly outperformed methods relying only on the RGB channels, all of which were based on deep neural networks (DNN's).
The PPF-based methods match pairs of oriented 3D points between the point cloud\footnote{The point cloud is calculated from the depth channel and known camera parameters.} of the test scene and the 3D object model, and aggregate the matches via a voting scheme. As each pair is described by only the distance and relative orientation of the two points, PPF-based methods can be effectively trained directly on the 3D object models, without the need to synthesize any training images. In contrast, DNN-based methods require large amounts of annotated training images, which have been typically obtained by OpenGL rendering of the 3D object models on top of
random backgrounds~\cite{kehl2017ssd,rad2017bb8,hinterstoisser2017pre,dwibedi2017cut}. However, as suggested in~\cite{hodan2019photorealistic}, the evident domain gap between these ``render\;\&\;paste'' training images and real test images presumably limits the potential of the DNN-based methods.

\input{fig_pbr_examples}

To reduce the gap between the synthetic and real domains and thus to bring fresh air to the DNN-based methods, we have created BlenderProc4BOP~\cite{denninger2019blenderproc,denninger2020blenderproc}, an open-source and light-weight physically-based renderer (PBR).
%and procedural data generator
Furthermore, to reduce the entry barrier of the challenge and to standardize the training set, the participants were provided with 350K pre-rendered PBR images (Fig.~\ref{fig:pbr_examples}).

In 2020, the DNN-based methods have finally caught up with the PPF-based methods -- five methods outperformed Vidal-Sensors18~\cite{vidal2018method}, the PPF-based winner from 2017 and 2019. Three of the top five methods, including the top-performing one, are single-view variants of CosyPose, a DNN-based method by Labb{\'e}~\etal~\cite{labbe2020cosypose}.
Strong data augmentation, similar to~\cite{sundermeyer2019augmented}, was identified as one of the key ingredients of this method. The second is a hybrid DNN+PPF method by K\"onig and Drost~\cite{koenig2020hybrid}, and the fourth is Pix2Pose, a DNN-based method by Park \etal~\cite{park2019pix2pose}. The first two methods used RGB-D image channels, while the third method achieved strong results with RGB channels only.

Methods achieved noticeably higher accuracy scores when trained on PBR training images than when trained on ``render\;\&\;paste'' images. Although adding real training images yielded even higher scores, competitive results were achieved with PBR images only -- out of the 26 evaluated methods, the fifth was trained only on PBR images. Interestingly, the increased photorealism from PBR images led to clear improvements of also the CosyPose method, despite the strong data augmentation which this method applies to the training images.

The rest of the paper is organized as follows. Section~\ref{sec:methodology} defines the evaluation methodology, Section~\ref{sec:datasets} introduces datasets and the implemented approach to synthesize photorealistic training images, Section~\ref{sec:evaluation} describes the experimental setup and analyzes the results, Section~\ref{sec:awards} presents the awards of the BOP Challenge 2020, and Section~\ref{sec:conclusion} concludes the paper. A discussion on the choices made when defining the evaluation methodology is provided in the supplement.

\section{Evaluation Methodology} \label{sec:methodology}

The evaluation methodology detailed in this section defines the challenge task, functions to measure the error of a 6D pose estimate, and calculation of the accuracy score used to compare the evaluated methods.
The BOP Challenge 2020 follows the same evaluation methodology as the BOP Challenge 2019 -- the scores have not been saturated and following the same methodology allowed using results from 2019 as baselines in 2020.

\subsection{Task Definition}

Methods are evaluated on the task of 6D localization of a \textbf{v}arying number of \textbf{i}nstances of a \textbf{v}arying number of \textbf{o}bjects from a single RGB-D image. This variant of the 6D object localization task is referred to as ViVo and defined as:

\customparagraph{Training input:}
For each object with an index $o \in \{1, \dots, k\}$, a method is given a 3D mesh model of the object (typically with a color texture) and a set of synthetic or real RGB-D images showing instances of the object in known 6D poses. The method may use any of the image channels.

\customparagraph{Test input:}
The method is provided with an image $I$ and a list $L = [(o_1, n_1),$ $\dots,$ $(o_m, n_m)]$, where $n_i$ is the number of instances of the object $o_i$ present in $I$.

\customparagraph{Test output:} The method produces a list $E = [E_1, \dots, E_m]$, where $E_i$ is a list of $n_i$ pose estimates for instances of the object $o_i$. Each estimate is given by a $3\times3$ rotation matrix $\mathbf{R}$, a $3\times1$ translation vector $\mathbf{t}$, and a confidence score~$s$. The matrix $\textbf{P} = [\mathbf{R} | \mathbf{t}]$ defines a rigid transformation from the 3D coordinate system of the object model to the 3D coordinate system of the camera.

\customparagraph{}Note that in the first challenge from 2017~\cite{hodan2017sixd,hodan2018bop}, methods were evaluated on a simpler variant of the 6D object localization task -- the goal was to estimate the 6D pose of a \textbf{s}ingle \textbf{i}nstance of a \textbf{s}ingle \textbf{o}bject (this variant is referred to as SiSo). If multiple instances of the same object model were visible in the image, then the pose of an arbitrary instance may have been reported. In 2017, the simpler SiSo variant was chosen because it allowed to evaluate all relevant methods out of the box. Since then, the state of the art has advanced and we have moved to the more challenging ViVo variant.

\subsection{Pose-Error Functions}

The error of an estimated pose $\hat{\textbf{P}}$ with respect to the ground-truth pose $\bar{\textbf{P}}$ of an object model $M$ is measured by three pose-error functions. The functions are defined below and discussed in more detail in the supplement.

\customparagraph{VSD (Visible Surface Discrepancy):}
\begin{equation}
	e_\mathrm{VSD}\big(\hat{D}, \bar{D}, \hat{V}, \bar{V}, \tau\big) =
	\mathrm{avg}_{p \in \hat{V} \cup \bar{V}}
	\begin{cases}
		0 & \text{if} \; p \in \hat{V} \cap \bar{V} \, \wedge \, \big|\hat{D}(p) -
		\bar{D}(p)\big| < \tau \\
		1 & \text{otherwise}
	\end{cases}
\end{equation}
\noindent The symbols $\hat{D}$ and $\bar{D}$ denote distance maps\footnote{A distance map stores at a pixel~$p$ the distance from the camera center to a 3D point $\mathbf{x}_p$ that projects to $p$. It can be readily computed from the depth map which stores at $p$ the $Z$ coordinate of $\mathbf{x}_p$ and which is a typical output of Kinect-like sensors.} obtained by rendering the object model $M$ in the estimated pose $\hat{\textbf{P}}$ and the ground-truth pose $\bar{\textbf{P}}$ respectively. These distance maps are compared with the distance map $D_I$ of the test image $I$ to obtain the visibility masks $\hat{V}$ and $\bar{V}$, \ie sets of pixels where the model $M$ is visible in the image $I$. The parameter $\tau$ is a misalignment tolerance.

Compared to~\cite{hodan2018bop,hodan2016evaluation}, estimation of the visibility masks has been modified -- an object is now considered visible at pixels with no depth measurements.
This modification allows evaluating poses of glossy objects from the ITODD dataset~\cite{drost2017introducing} whose surface is not always captured in the depth image channel.

VSD treats poses that are indistinguishable in shape (color is not considered) as equivalent by measuring the misalignment of only the visible part of the object surface. See Sec.~2.2 of~\cite{hodan2018bop} and the supplement of this paper for details.

\customparagraph{MSSD (Maximum Symmetry-Aware Surface Distance):}
\begin{equation}
	e_{\text{MSSD}}\big(\hat{\mathbf{P}}, \bar{\mathbf{P}}, S_M, V_M\big) = \text{min}_{\textbf{S} \in S_M} \text{max}_{\textbf{x}
		\in V_M}
	\big\Vert \hat{\textbf{P}}\textbf{x} - \bar{\textbf{P}}\textbf{S}\textbf{x}
	\big\Vert_2
\end{equation}
\noindent The set $S_M$ contains global symmetry transformations of the object model $M$, identified as described in Sec.~\ref{sec:symmetries}, and $V_M$ is a set of the model vertices.

The maximum distance between the model vertices is relevant for robotic manipulation, where the maximum surface deviation strongly indicates the chance of a successful grasp.
Moreover, compared to the average distance used in pose-error functions ADD and ADI~\cite{hodan2016evaluation,hinterstoisser2012accv}, the maximum distance is less dependent on the geometry of the object model and the sampling density of its surface.

\customparagraph{MSPD (Maximum Symmetry-Aware Projection Distance):}
\begin{equation}
	e_{\text{MSPD}}\big(\hat{\mathbf{P}}, \bar{\mathbf{P}}, S_M, V_M\big) = \text{min}_{\textbf{S} \in S_M} \text{max}_{\textbf{x}
		\in V_M}
	\big\Vert \text{proj}\big( \hat{\textbf{P}}\textbf{x} \big) - \text{proj}\big(
	\bar{\textbf{P}}\textbf{S}\textbf{x} \big) \big\Vert_2
\end{equation}
\noindent The function $\text{proj}(\cdot)$ is the 2D projection (the result is in pixels) and the meaning of the other symbols is as in MSSD.

Compared to the pose-error function from~\cite{brachmann2014learning}, MSPD considers global object symmetries and replaces the average by the maximum distance to increase robustness against the geometry and sampling of the object model.
Since MSPD does not evaluate the alignment along the optical ($Z$) axis and measures only the perceivable discrepancy, it is relevant for augmented reality applications and suitable for evaluating RGB-only methods, for which estimating the alignment along the optical axis is more challenging.

\subsection{Identifying Global Object Symmetries} \label{sec:symmetries}

The set of global symmetry transformations of an object model $M$, which is used in the calculation of MSSD and MSPD, is identified in two steps. Firstly, a set of candidate symmetry transformations is defined as $S'_M = \{\textbf{S}: h(V_M, \textbf{S}V_M) < \varepsilon \}$, where $h$ is the Hausdorff distance calculated between the vertices $V_M$ of the object model $M$ in the canonical and the transformed pose. The allowed deviation is bounded by $\varepsilon = \text{max}(15\,\text{mm}, 0.1d)$, where $d$ is the diameter of the object model $M$ (the largest distance between any pair of vertices) and the truncation at $15\,$mm avoids breaking the symmetries by too small details. Secondly, the final set of symmetry transformations $S_M$ is defined as a subset of $S'_M$ which consists of those symmetry transformations that cannot be resolved by the model texture (decided subjectively by the organizers of the challenge).

The set $S_M$ covers both discrete and continuous global rotational symmetries. The continuous rotational symmetries are discretized such as the vertex which is the furthest from the axis of symmetry travels not more than $1\%$ of the object diameter between two consecutive rotations.

\subsection{Accuracy Score} \label{sec:accuracy_score}

An estimated pose is considered correct with respect to a pose-error function $e$, if $e < \theta_e$, where $e \in \{e_{\text{VSD}}, e_{\text{MSSD}}, e_{\text{MSPD}}\}$ and $\theta_e$ is a threshold of correctness.

The fraction of annotated object instances for which a correct pose is estimated is referred to as Recall. The Average Recall with respect to a function~$e$, denoted as $\text{AR}_e$, is defined as the average of Recall rates calculated for multiple settings of the threshold $\theta_e$, and also for multiple settings of the misalignment tolerance $\tau$ in the case of $e_{\text{VSD}}$.
In particular, $\text{AR}_\text{VSD}$ is the average of Recall rates calculated for $\tau$ ranging from $5\%$ to $50\%$ of the object diameter with a step of $5\%$, and for $\theta_{\text{VSD}}$ ranging from $0.05$ to $0.5$ with a step of $0.05$.
$\text{AR}_\text{MSSD}$ is the average of Recall rates calculated for $\theta_{\text{MSSD}}$ ranging from $5\%$ to $50\%$ of the object diameter with a step of $5\%$.
Finally, $\text{AR}_\text{MSPD}$ is the average of Recall rates calculated for $\theta_{\text{MSPD}}$ ranging from $5r$ to $50r$ with a step of $5r$, where $r = w/640$ and $w$ is the image width in pixels.

The accuracy of a method on a dataset $D$ is measured by $\text{AR}_D = (\text{AR}_{\text{VSD}} + \text{AR}_{\text{MSSD}} + \text{AR}_{\text{MSPD}}) / 3$. The overall accuracy on the core datasets is then measured by $\text{AR}_{\text{Core}}$ defined as the average of the per-dataset $\text{AR}_D$ scores. In this way, each dataset is treated as a separate sub-challenge which avoids $\text{AR}_{\text{Core}}$ being dominated by larger datasets.

\section{Datasets} \label{sec:datasets}

BOP currently includes eleven datasets in a unified format, detailed in Tab.~\ref{tab:dataset_params}, seven of which were selected as core datasets. A method had to be evaluated on all core datasets to be considered for the main challenge awards (Sec.~\ref{sec:awards}).

\input{fig_datasets}

\subsection{Content of Datasets}

Each dataset is provided in a unified format and includes 3D object models and training and test RGB-D images annotated with ground-truth 6D object poses.
The HB and ITODD datasets include also validation images -- in this case, the ground-truth poses are publicly available only for the validation images, not for the test images.
The object models were created manually or using KinectFusion-like systems for 3D surface reconstruction~\cite{newcombe2011kinectfusion}.
The seven core datasets include photorealistic training images described in Sec.~\ref{sec:synthesis}. Datasets T-LESS, TUD-L, and YCB-V include real training images, and most datasets include also training images obtained by OpenGL rendering of the 3D object models on a black background.
The test images were captured in scenes with graded complexity, often with clutter and occlusion.
The datasets can be downloaded from: \texttt{\href{http://bop.felk.cvut.cz/datasets}{bop.felk.cvut.cz/datasets}}.

\subsection{Photorealistic Training Images} \label{sec:synthesis}

In the BOP Challenge 2020, the participants were provided with 50K photorealistic training images for each of the seven core datasets. The images were generated and automatically annotated by BlenderProc4BOP~\cite{denninger2019blenderproc,denninger2020blenderproc}, an open-source and light-weight physically-based renderer of procedurally generated scenes.

Physically-based rendering (PBR) accurately simulates the flow of light energy in the scene by ray tracing. This naturally accounts for complex illumination effects such as scattering, refraction and reflection, including diffuse and specular interreflection between the scene elements~\cite{pharr2016physically}.
The rendered images look realistic and are often difficult to differentiate from real photographs. Rendering techniques based on rasterization, \eg OpenGL, can approximate the complex effects in an ad hoc way through custom shaders, but the approximations cause physically incorrect artifacts that are difficult to eliminate~\cite{marschner2015fundamentals}.

BlenderProc4BOP implements a PBR synthesis approach similar to~\cite{hodan2019photorealistic}. However, to improve efficiency, objects are not rendered in 3D models of complete indoor scenes but inside an empty cube, with objects from other BOP datasets serving as distractors.
To achieve a rich spectrum of the generated images, a random PBR material from the CC0 Textures library~\cite{demes2020textures} is assigned to the walls of the cube, and light with a random intensity and color is emitted from the room ceiling and from a randomly positioned point source.
The number of rays traced per image pixel is set to $50$ and the Intel Open Image Denoiser~\cite{inteldenoise} is applied to reduce noise in the rendered image.
This setup keeps the computational cost low -- the full generation of one $640\times480$ RGB-D image takes \textbf{1--3 seconds} on a standard desktop computer with a modern GPU. A set of 50K images can be therefore rendered on 5 GPU's overnight.

Instead of trying to accurately model the object materials, properties such as specularity, roughness and metallicness are randomized. Such physically plausible domain randomization is important since objects in the challenge as well as in real-world scenarios are typically not modeled perfectly. Realistic object poses are achieved by dropping the 3D object models on the ground plane of the cube using the PyBullet physics engine integrated in Blender~\cite{blender}. This allows to create dense but shallow piles of objects that introduce various levels of occlusion. 
Since test images from the LM dataset show the objects always standing upright, the objects from LM are not dropped but instead densely placed on the ground plane in upright poses using automated collision checks.

Each object arrangement is rendered from $25$ random camera poses. Instead of fitting all objects within the camera frustum, each camera is pointed at a randomly selected object close to the center, which allows generating more diverse camera poses. The azimuth angles, elevation angles, and distances of the cameras are uniformly sampled from ranges determined by the ground-truth 6D object poses from the test images. In-plane rotation angles are generated randomly.

The generated data (object poses, camera intrinsics, RGB and depth) is saved in the BOP format, allowing to interface with utilities from the BOP toolkit~\cite{boptoolkit}. 
Configuration files to reproduce or modify the generation process are provided.\footnote{\texttt{\href{https://github.com/DLR-RM/BlenderProc/blob/master/README_BlenderProc4BOP.md}{github.com/DLR-RM/BlenderProc/blob/master/README\_BlenderProc4BOP.md}}}

\section{Evaluation} \label{sec:evaluation}

This section presents the results of the BOP Challenge 2020, compares them with the results from 2019, and analyzes the effectiveness of PBR training images.

%This section describes the experimental setup, presents the results of the BOP Challenge 2020, and analyzes the effectiveness of PBR training images.

\subsection{Experimental Setup}

Participants of the BOP Challenge 2020 were submitting the results of their methods to the online evaluation system at \texttt{\href{http://bop.felk.cvut.cz/}{bop.felk.cvut.cz}} from June 5th, 2020, until the deadline on August 19th, 2020. The methods were evaluated on the ViVo variant of the 6D object localization task as described in Sec.~\ref{sec:methodology}. The evaluation script is publicly available in the BOP toolkit~\cite{boptoolkit}.

A method had to use a fixed set of hyper-parameters across all objects and datasets. For training, a method may have used the provided object models and training images, and rendered extra training images using the object models. However, not a single pixel of test images may have been used for training, nor the individual ground-truth poses or object masks provided for the test images.
Ranges of the azimuth and elevation camera angles, and a range of the camera-object distances determined by the ground-truth poses from test images is the only information about the test set that may have been used for training.
%The range of all ground-truth poses in the test images, is the only information about the test set that may have been used for training.

Only subsets of test images were used to remove redundancies and speed up the evaluation, and only object instances for which at least $10\%$ of the projected surface area is visible were to be localized in the selected test images.
%A list of selected test images and object instances for which the 6D pose is to be estimated is on the project website.
%visible from at least $10\%$ are included in the list.

\input{tab_results}

\subsection{Results} \label{sec:results}

In total, 26 methods were evaluated on all seven core datasets. Results of 11 methods were submitted to the BOP Challenge 2019 and results of 15 methods to the BOP Challenge 2020 (column ``Year'' in Tab.~\ref{tab:results}).

In 2020, methods based on deep neural networks (DNN's) have finally caught up with methods based on point pair features (PPF's)~\cite{drost2010model} -- five methods from 2020 outperformed Vidal-Sensors18~\cite{vidal2018method}, the PPF-based winner of the first two challenges from 2017 and 2019 (columns ``PPF'' and ``DNN'' in Tab.~\ref{tab:results}).
Almost all participating DNN-based methods applied neural networks only to the RGB image channels and many of these methods used the depth channel for ICP refinement at test time (columns ``Train'', ``Test'', and ``Refine'').
Only PointVoteNet2~\cite{hagelskjaer2019pointposenet} applied a neural network also to the depth channel.
It is noteworthy that the overall third method does not use the depth channel at all.

Three of the top five methods, including the top-performing one, are single-view variants of the CosyPose method by Labb{\'e}~\etal~\cite{labbe2020cosypose}.
This method first predicts 2D bounding boxes of the objects using Mask R-CNN~\cite{he2017mask}, and then applies to each box a DNN model for coarse pose estimation followed by a DNN model for iterative refinement.
The top variant of CosyPose, with the $\text{AR}_{\text{Core}}$ score of $69.8\%$, additionally applies a depth-based ICP refinement which improves the score by $6.1\%$ (method \#1 \vs \#3 in Tab.~\ref{tab:results}).
One of the key ingredients of CosyPose is a strong data augmentation technique similar to~\cite{sundermeyer2018implicit}. As reported in~\cite{labbe2020cosypose}, using the augmentation for training the pose estimation models improves the
accuracy on T-LESS from $37.0\%$ to $63.8\%$. Access to a GPU cluster is also crucial as training of a model takes $\mytilde$10 hours on 32 GPU's.

The second is a hybrid method by K\"onig and Drost~\cite{koenig2020hybrid} with $\text{AR}_{\text{Core}}$ of $63.9\%$. This method first predicts object instance masks by RetinaMask~\cite{fu2019retinamask} or Mask R-CNN~\cite{he2017mask}, whichever performs better on the validation set. Then, for each mask, the method selects the corresponding part of the 3D point cloud of the test scene, and estimates the object pose using the point pair features~\cite{drost2010model}. The method is noticeably faster than the top-performing
CosyPose variant, mainly thanks to a highly optimized implementation of ICP from HALCON~\cite{halcon}.

Another method which outperformed Vidal-Sensors18 is Pix2Pose by Park \etal~\cite{park2019pix2pose} with $\text{AR}_{\text{Core}}$ of $59.1\%$. This method predicts 2D-3D correspondences between densely sampled image pixels and the 3D object model, solves for the poses using the P\emph{n}P-RANSAC algorithm, and refines the poses with a depth-based ICP algorithm. The ICP refinement is crucial for this method as it improves the $\text{AR}_{\text{Core}}$ score by absolute $24.9\%$ and teleports the method from the 22nd to the 4th place. The importance of a refinement stage has been demonstrated also by other methods -- top nine methods applied ICP or an RGB-based refiner, similar to DeepIM~\cite{li2018deepim} (column ``Refine'' in Tab.~\ref{tab:results}).

Training a special DNN model per object has been a common practise in the field, followed also by most participants of the challenge. However, the CosyPose and K\"onig-Hybrid-DL-PointPairs methods have shown that a single DNN model can be effectively shared among multiple objects (column ``DNN'' in Tab.~\ref{tab:results}). CosyPose trains three models per dataset -- one for detection, one for coarse pose estimation, and one for iterative pose refinement, whereas K\"onig-Hybrid-DL-PointPairs trains only one model for instance segmentation.

\subsection{The Effectiveness of Photorealistic Training Images} \label{sec:rgb_types}

In 2020, most DNN-based methods were trained either only on the photorealistic (PBR) training images, or also on real training images which are available in datasets T-LESS, TUD-L, and YCB-V (column ``Train type'' in Tab.~\ref{tab:results}).\footnote{Method \#2 used also synthetic training images obtained by cropping the objects from real validation images in the case of HB and ITODD and from OpenGL-rendered images in the case of other datasets, and pasting the cropped objects on images from the Microsoft COCO dataset~\cite{lin2014microsoft}. Method \#24 used PBR and real images for training Mask R-CNN~\cite{he2017mask} and OpenGL images for training a single Multi-path encoder. Two of the CosyPose variants (\#1 and \#3) also added the ``render\;\&\;paste'' synthetic images provided in the original YCB-V dataset, but these images were later found to have no effect on the accuracy score.}
Although adding real training images yields higher scores (compare scores of methods \#3 and \#5 or \#10 and \#14 on T-LESS, TUD-L, and YCB-V in Tab.~\ref{tab:results}), competitive results can be achieved with PBR images only, as demonstrated by the overall fifth PBR-only variant of the CosyPose method. This is an important result considering that PBR-only training does not require any human effort for capturing and annotating real training images.

\input{tab_cosypose_data}

The PBR training images yield a noticeable improvement over the ``render\;\& paste''
synthetic images obtained by OpenGL rendering of the 3D object models on real photographs.
For example, the CDPN method with the same hyper-parameter settings improved by absolute $20.2\%$ on HB, by $19.5\%$ on LM-O, and by $7\%$ on IC-BIN when trained on 50K PBR images per dataset \vs 10K ``render\;\&\;paste'' images per object (compare methods \#13 and \#20 in Tab.~\ref{tab:results}).
As shown in Tab.~\ref{tab:cosy_data_res}, the CosyPose method improved by a significant $57.9\%$ (from $6.1\%$ to $64.0\%$) on T-LESS, by $19.0\%$ on TUD-L, and by $30.9\%$ on YCB-V when trained on 50K PBR images per dataset \vs 50K ``render\;\&\;paste v1'' images per dataset.
The ``render\;\&\;paste v1'' images used for training CosyPose were obtained by imitating the PBR images, \ie~the 3D object models were rendered in the same poses as in the PBR images and pasted on real backgrounds.

As an additional experiment, we have trained the CosyPose method on another variant of the ``render\;\&\;paste'' images, generated as in~\cite{labbe2020cosypose} and referred to as ``render\;\&\;paste v2''. The main differences compared to the ``render\;\&\;paste v1'' variant described in the previous paragraph are: (a) the CAD models of T-LESS objects were assigned a random surface texture instead of a random gray value, (b) the background was assigned a real photograph in 30\% images and a synthetic texture in 70\% images, and (c) 1M instead of 50K images were generated.
As shown in Tab.~\ref{tab:cosy_data_res}, ``render\;\&\;paste v2'' images yield a noticeable improvement of $39.2\%$ over ``render\;\&\;paste v1'' on T-LESS, but no improvement on TUD-L ($-7.1\%$) and YCB-V ($-0.8\%$).
This may suggest that randomizing the surface texture of the texture-less CAD models of T-LESS objects improves the generalization of the network by forcing the network to focus more on shape than on lower-level patterns, as found in~\cite{geirhos2018imagenet}.
When generating the PBR images, which yield the highest accuracy on T-LESS, the CAD models were assigned a random gray value, as in ``render\;\&\;paste v1'', but the effect of randomizing the surface texture may have been achieved by randomizing the PBR material (Sec.~\ref{sec:synthesis}) -- further investigation is needed to clearly answer these questions.
The importance of both the objects and the background being synthetic, as suggested in~\cite{hinterstoisser2019annotation}, has not been confirmed in this experiment -- ``render\;\&\;paste v1'' images with only real backgrounds achieved higher scores than ``render\;\&\;paste v2'' images on TUD-L and YCB-V.
However, the first ten convolutional layers of Mask R-CNN (``conv1'' and ``conv2\_x'' of ResNet-50~\cite{he2016deep}) used for object detection in CosyPose were pre-trained on Microsoft COCO~\cite{lin2014microsoft} but not fine-tuned, whereas all layers were fine-tuned in~\cite{hinterstoisser2019annotation}.
The benefit of having 1M \vs 50K images is indecisive since 50K PBR images were sufficient to achieve high scores.

Both types of ``render\;\&\;paste'' images are far inferior compared to the PBR images, which yield an average improvement of $35.9\%$ over ``render\;\&\;paste v1'' and $25.5\%$ over ``render\;\&\;paste v2'' images (Tab.~\ref{tab:cosy_data_res}).
Interestingly, the increased photorealism brought by the PBR images is important despite the strong data augmentation that CosyPose applies to the training images. Since object poses in the PBR and ``render\;\&\;paste v1'' images are identical, the ray-tracing rendering technique, PBR materials and objects realistically embedded in synthetic environments seem to be the decisive factors for successful ``sim2real'' transfer~\cite{denninger2020blenderproc}.

We have also observed that the PBR images are more important for training DNN models for object detection/segmentation (\eg Mask R-CNN~\cite{he2017mask}) than for training DNN models for pose estimation from the detected regions (Tab.~\ref{tab:cosy_data_res}). In the case of CosyPose, if the detection model is trained on PBR images and the later two models for pose estimation are trained on the ``render\;\&\;paste v2'' instead of the PBR images, the accuracy drops moderately ($64.0\%$ to $60.0\%$ on T-LESS, $68.5\%$ to $58.9\%$ on TUD-L) or does not change much ($57.4\%$ \vs $58.5\%$ on YCB-V). However, if also the detection model is trained on the ``render\;\&\;paste v1'' or ``render\;\&\;paste v2'' images, the accuracy drops severely (the low accuracy achieved with ``render\;\&\;paste v1'' on T-LESS was discussed earlier).

\section{Awards} \label{sec:awards}

The following BOP Challenge 2020 awards were presented at the 6th Workshop on Recovering 6D Object Pose~\cite{hodan2020r6d} organized in conjunction with the ECCV 2020 conference. Results on the core datasets are in Tab.~\ref{tab:results} and results on the other datasets can be found on the project website.

\customparagraph{The Overall Best Method} (the top-performing method on the core datasets): CosyPose-ECCV20-Synt+Real-ICP by Yann Labb{\'e}, Justin Carpentier, Mathieu Aubry, and Josef Sivic~\cite{labbe2020cosypose}.

\customparagraph{The Best RGB-Only Method} (the top-performing RGB-only method on~the core datasets): CosyPose-ECCV20-Synt+Real by Yann Labb{\'e}, Justin Carpentier, Mathieu Aubry, and Josef Sivic~\cite{labbe2020cosypose}.

\customparagraph{The Best Fast Method} (the top-performing method on the core datasets with the average running time per image below 1s):
K\"onig-Hybrid-DL-PointPairs by Rebecca K\"onig and Bertram Drost~\cite{koenig2020hybrid}.

\customparagraph{The Best BlenderProc4BOP-Trained Method} (the top-performing method on the core datasets which was trained only with the provided BlenderProc4BOP images):
CosyPose-ECCV20-PBR by Yann Labb{\'e}, Justin Carpentier, Mathieu Aubry, and Josef Sivic~\cite{labbe2020cosypose}.

\customparagraph{The Best Single-Model Method} (the top-performing method on the core datasets which uses a single machine learning model, typically a neural network, per dataset): CosyPose-ECCV20-Synt+Real-ICP by Yann Labb{\'e}, Justin Carpentier, Mathieu Aubry, and Josef Sivic~\cite{labbe2020cosypose}.

\customparagraph{The Best Open-Source Method} (the top-performing method on the core datasets whose source code is publicly available): CosyPose-ECCV20-Synt+Real-ICP by Yann Labb{\'e}, Justin Carpentier, Mathieu Aubry, and Josef Sivic~\cite{labbe2020cosypose}.

\customparagraph{The Best Method on Datasets LM-O,\;TUD-L,\;IC-BIN,\;and\;YCB-V:}   CosyPose-ECCV20-Synt+Real-ICP by Yann Labb{\'e}, Justin Carpentier, Mathieu Aubry, and Josef Sivic~\cite{labbe2020cosypose}.

\customparagraph{The Best Method on Datasets ITODD and TYO-L:} Drost-CVPR10-Edges by Bertram Drost, Markus Ulrich, Nassir Navab, and Slobodan Ilic~\cite{drost2010model}.

\customparagraph{The Best Method on Dataset LM:} DPODv2 (synthetic train data, RGB\ +\ D Kabsch) by Sergey Zakharov, Ivan Shugurov, and Slobodan Ilic~\cite{zakharov2019dpod}.

\customparagraph{The Best Method on Dataset T-LESS:} CosyPose-ECCV20-Synt+Real by Yann Labb{\'e}, Justin Carpentier, Mathieu Aubry, and Josef Sivic~\cite{labbe2020cosypose}.

\customparagraph{The Best Method on Dataset HB:} CDPNv2\_BOP20 (RGB-only) by Zhigang Li, Gu Wang, and Xiangyang Ji~\cite{li2019cdpn}.

\customparagraph{The Best Method on Dataset RU-APC:} Pix2Pose-BOP19\_w/ICP-ICCV19 by Kiru Park, Timothy Patten, and Markus Vincze~\cite{park2019pix2pose}.

\customparagraph{The Best Method on Dataset IC-MI:}  Drost-CVPR10-3D-Only by Bertram Drost, Markus Ulrich, Nassir Navab, and Slobodan Ilic~\cite{drost2010model}.

\section{Conclusions} \label{sec:conclusion}

In 2020, methods based on neural networks have finally caught up with methods based on point pair features, which were dominating previous editions of the challenge. Although the top-performing methods rely on RGB-D image channels, strong results have been achieved with RGB channels only.
The challenge results and additional experiments with the top-performing CosyPose method~\cite{labbe2020cosypose} have shown the importance of PBR training images and of strong data augmentation for successful ``sim2real'' transfer.
The scores have not been saturated and we are already looking forward to the insights from the next challenge.

\vspace{1.0em}
\begin{spacing}{0.85}
\begin{footnotesize}
\noindent
This research was supported by CTU student grant (SGS OHK3-019/20), Research Center for Informatics (CZ.02.1.01/0.0/0.0/16\_019/0000765 funded by OP VVV), and HPC resources from GENCI-IDRIS (grant 011011181).
\end{footnotesize}
\end{spacing}
\normalsize

\bibliographystyle{splncs04}
\bibliography{references}

\appendix

\section{Discussion on the Evaluation Methodology}

\subsection{6D Object Localization \vs 6D Object Detection}

Prior information about the object presence in the input image distinguishes two 6D object pose estimation tasks: 6D object localization, where the identifiers of present object instances are provided for each image, and 6D object detection, where no prior information is provided~\cite{hodan2016evaluation}.

The aspect which is evaluated on the 6D object detection but not on the 6D object localization task is the capability of the method to calibrate the predicted confidence scores across all object classes. For example, a score of $0.5$ for a cat should represent the same level of confidence as a score of $0.5$ for a duck. This calibration is important for achieving good performance \wrt the precision/recall curve which is typically used for evaluating detection. The 6D object localization task still requires the method to sort the hypotheses, although only within the same object class -- the method needs to output the top $n$ pose estimates for an object class which are evaluated against $n$ ground-truth poses of that class.

In BOP, methods have been so far evaluated on the 6D object localization task for two reasons. First, the accuracy scores on this simpler task are still far from being saturated. Second, the 6D object detection task requires computationally expensive evaluation as many more hypotheses need to be evaluated to calculate the precision/recall curve. Calculating the 6D pose errors is more expensive than, \eg, calculating the intersection over union of 2D bounding boxes (used to evaluate 2D object detection).

\subsection{The Choice of Pose-Error Functions} \label{sec:error_discussion}

The object pose may be ambiguous, \ie, there may be multiple poses that are consistent with the image. This is caused by the existence of multiple fits of the visible part of the object surface to the 3D object model. The visible part is determined by self-occlusion and occlusion by other objects and the multiple surface fits are induced by global or partial object symmetries. As a consequence, there may be (infinitely) many indistinguishable 6D poses which should be treated as equivalent, but explicitly enumerating all of these poses is often difficult.

The most widely used pose-error functions have been ADD/ADI~\cite{hodan2016evaluation,hinterstoisser2012accv}, where the error is calculated as the average distance from vertices of the object model in the ground-truth pose to vertices of the model in the estimated pose. The distance is measured between corresponding vertices if all views of the object are distinguishable (ADD). Otherwise, for objects with indistinguishable views, the distance is measured between a vertex and its nearest neighbor in the 3D space, which may not necessarily be the corresponding vertex (ADI). ADI can yield unintuitively low errors even for poses that are distinguishable. Objects evaluated with ADI therefore tend to have low pose errors although the estimated poses might not be visually well aligned. Another limitation of ADD/ADI comes from a high dependence on the geometry of the object model and the sampling density of its surface -- the average distance is dominated by higher-frequency surface parts such as the thread of a fuse. The maximum distance used in MSSD and MSPD is less dependent on the geometry and sampling of the object model.

MSSD is relevant for robotic grasping as it measures the error in the 3D space, and MSPD is relevant for augmented reality applications as it measures the error in the projective space. Both MSSD and MSPD can handle pose ambiguities due to global object symmetries. However, because both are calculated over the entire model surface, misalignments of invisible surface parts are penalized. This may not be desirable for applications such as robotic manipulation with suction cups where only the alignment of the visible part is relevant.
VSD is calculated only over the visible object part and therefore treats all poses that are consistent with the image as equivalent. VSD evaluates the alignment of the object shape but not of its color. This is because most of the object models currently included in BOP have baked shadows and reflections in their surface textures, which makes it difficult to robustly evaluate the color alignment.

As each of VSD, MSSD, and MSPD evaluates different qualitites of the pose estimates and each is relevant for a different target application, we use all three of these pose-error functions for the evaluation in BOP.

\end{document}

%% file: fig_pbr_examples.tex
\begin{figure}[t!]
\begin{center}

\begingroup
\renewcommand{\arraystretch}{0.9}

\begin{tabular}{ @{}c@{ } @{}c@{ } @{}c@{ } @{}c@{ } }

\multicolumn{4}{c}{Commonly used ``render\;\&\;paste'' synthetic training images} \vspace{0.5ex} \\

\includegraphics[width=0.243\columnwidth]{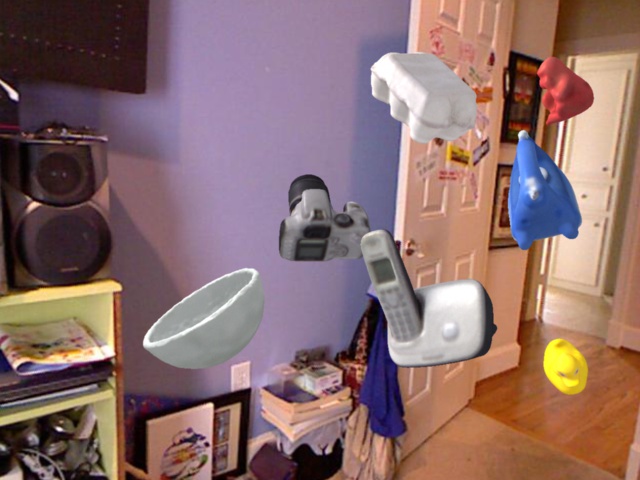} &
\includegraphics[width=0.243\columnwidth]{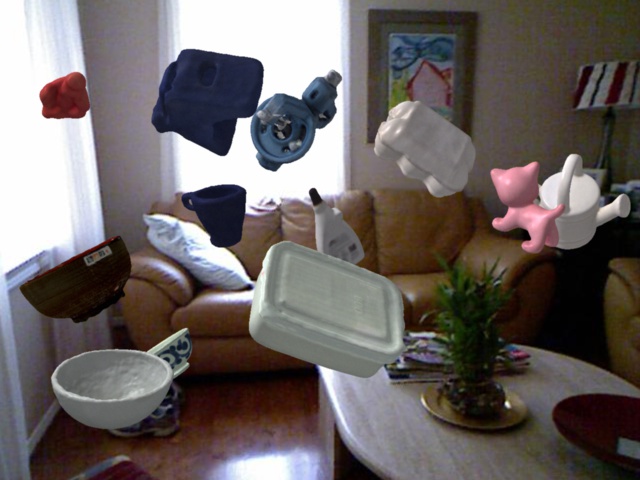} &
\includegraphics[width=0.243\columnwidth]{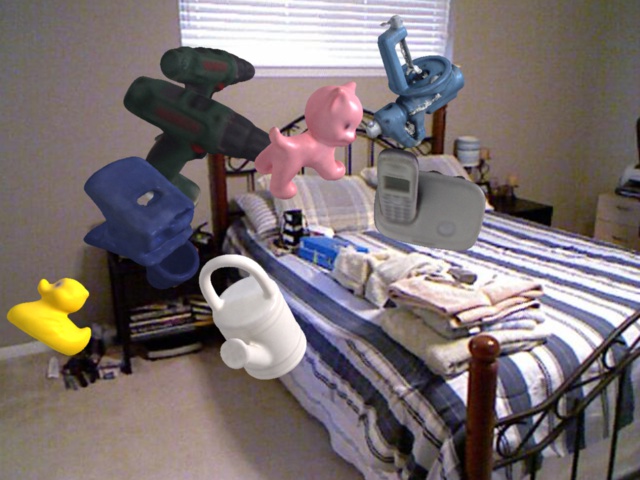} &
\includegraphics[width=0.243\columnwidth]{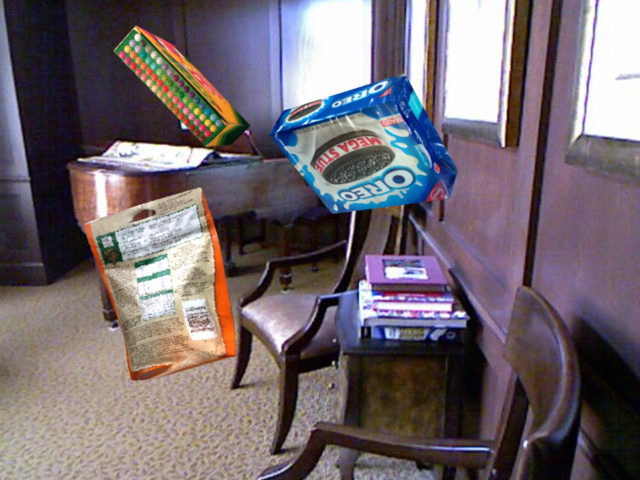} \\

\multicolumn{4}{c}{}\vspace{-2.0ex} \\
\multicolumn{4}{c}{Photorealistic training images rendered by BlenderProc4BOP~\cite{denninger2019blenderproc,denninger2020blenderproc}} \vspace{0.5ex} \\

\includegraphics[width=0.243\columnwidth]{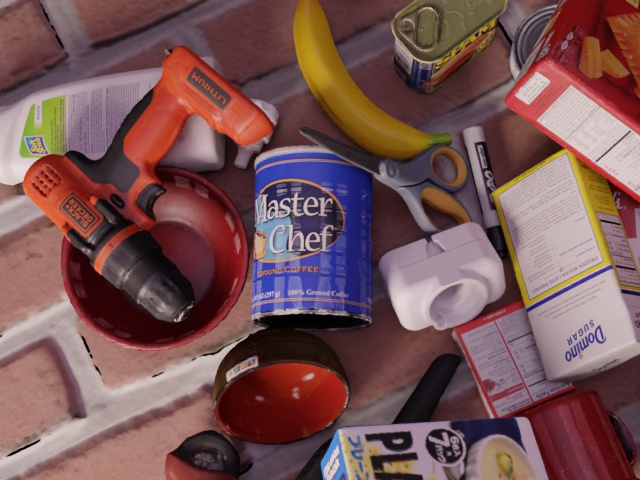} &
\includegraphics[width=0.243\columnwidth]{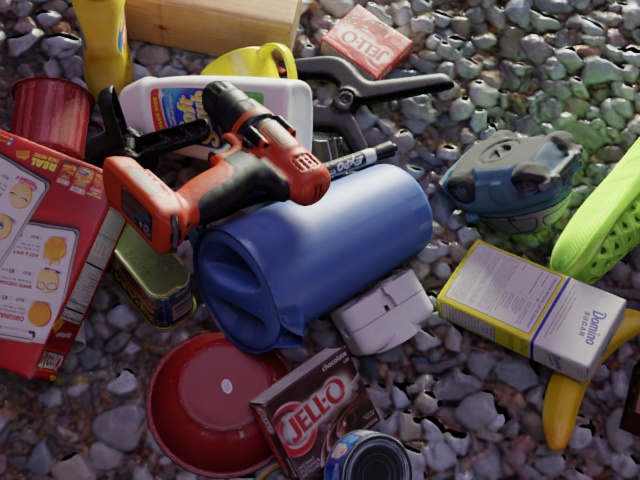} &
\includegraphics[width=0.243\columnwidth]{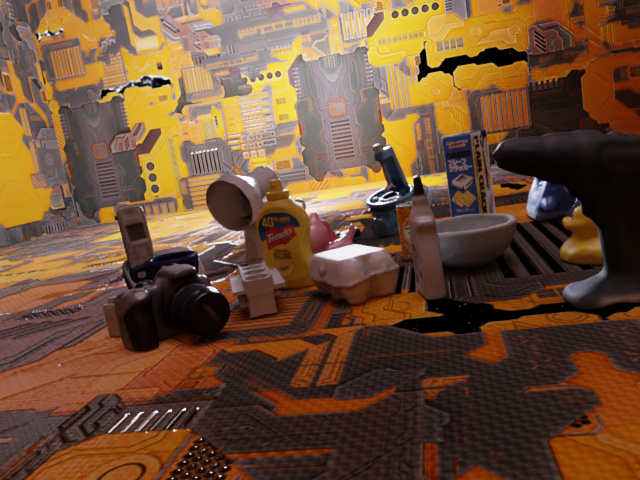} &
\includegraphics[width=0.243\columnwidth]{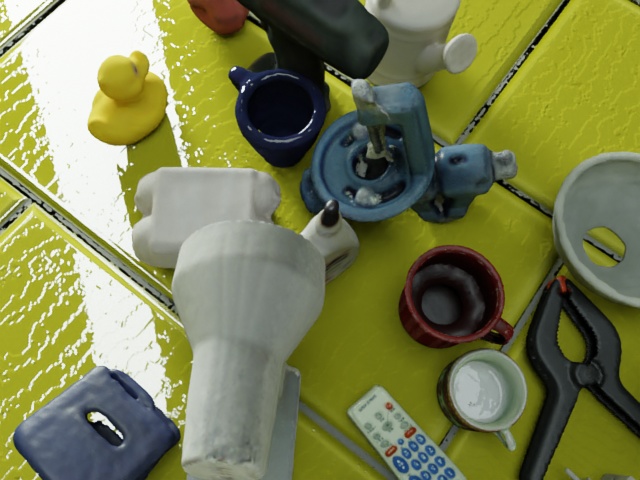} \\

\includegraphics[width=0.243\columnwidth]{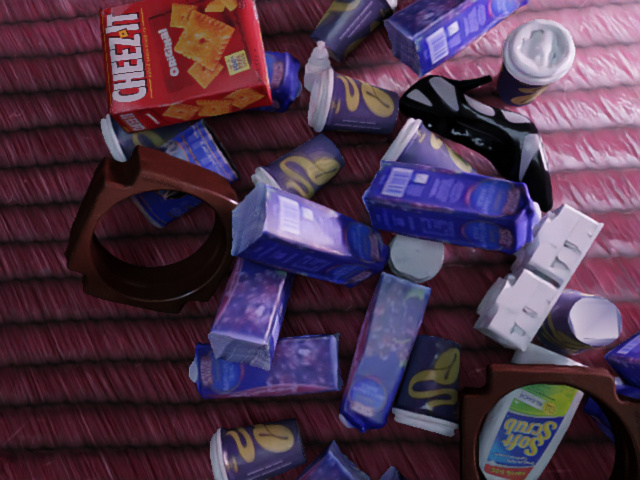} &
\includegraphics[width=0.243\columnwidth]{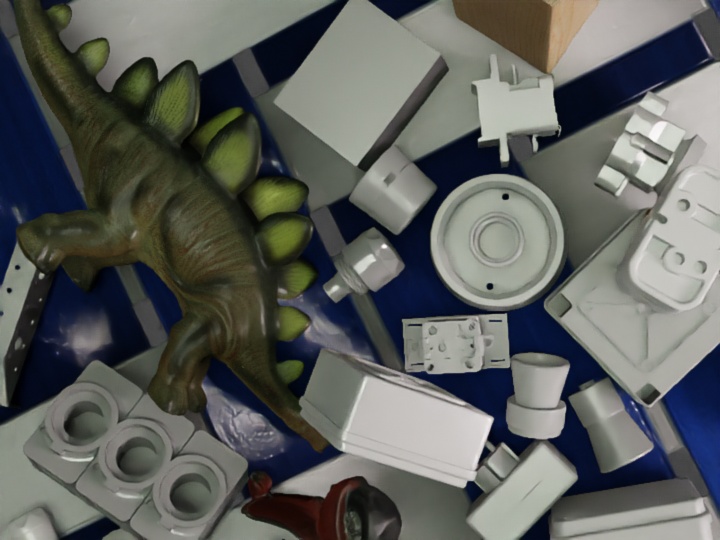} &
\includegraphics[width=0.243\columnwidth]{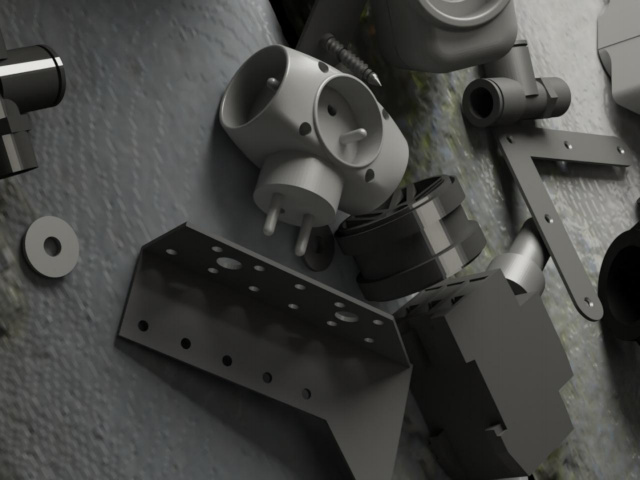} &
\includegraphics[width=0.243\columnwidth]{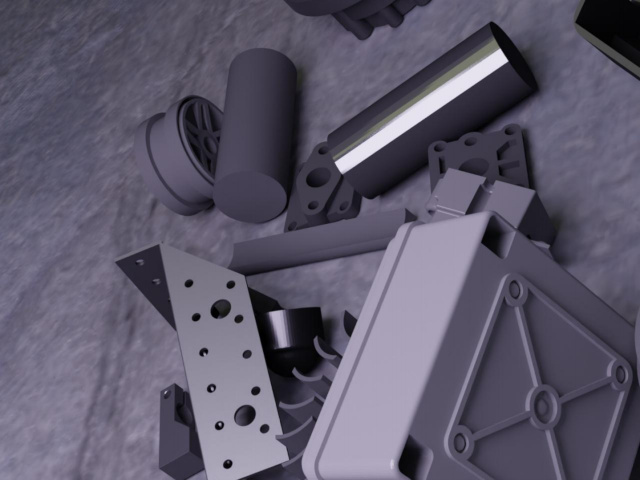} \\

\end{tabular}
\endgroup

\caption{\label{fig:pbr_examples} \textbf{Synthetic training images.} DNN-based methods for 6D object pose estimation have been commonly trained on ``render\;\&\;paste'' images synthesized by OpenGL rendering of 3D object models randomly positioned on top of random backgrounds. Instead, participants of the BOP Challenge 2020 were provided 350K photorealistic training images synthesized by ray tracing and showing the 3D object models in physically plausible poses inside a cube with a random PBR material (see Sec.~\ref{sec:synthesis}).}

\end{center}
\end{figure}

%% file: fig_datasets.tex
\begin{figure}[h!]
\begin{center}

\begingroup
\footnotesize
\begin{tabularx}{\textwidth}{ l *{11}{Y} }
	\toprule
	&
	&
	&
	\multicolumn{2}{c}{Train. im.} &
	\multicolumn{1}{c}{Val im.} &
	\multicolumn{2}{c}{Test im.} &
	\multicolumn{2}{c}{Test inst.} \\
	\cmidrule(l{2pt}r{2pt}){4-5} \cmidrule(l{2pt}r{2pt}){6-6} \cmidrule(l{2pt}r{2pt}){7-8} \cmidrule(l{2pt}r{2pt}){9-10}
	\multicolumn{1}{l}{Dataset} &
	\multicolumn{1}{c}{Core} &
	\multicolumn{1}{c}{Objects} &
	\multicolumn{1}{c}{Real} &
	\multicolumn{1}{c}{PBR} &
	\multicolumn{1}{c}{Real} &
	\multicolumn{1}{c}{All} &
	\multicolumn{1}{c}{Used} &
	\multicolumn{1}{c}{All} &
	\multicolumn{1}{c}{Used} \\
	\midrule
	LM \cite{hinterstoisser2012accv} & & 15 & $\phantom{00000}$-- & 50000 & $\phantom{000}$-- & 18273 & 3000 & 18273 & 3000 \\
	LM-O \cite{brachmann2014learning} & $\ast$ & $\phantom{0}$8 & $\phantom{00000}$-- & 50000 & $\phantom{000}$-- & $\phantom{0}$1214 & $\phantom{0}$200 & $\phantom{0}$9038 & 1445 \\
	T-LESS \cite{hodan2017tless} & $\ast$ & 30 & $\phantom{0}$37584 & 50000 & $\phantom{000}$-- & 10080 & 1000 & 67308 & 6423 \\
	ITODD \cite{drost2017introducing} & $\ast$ & 28 & $\phantom{00000}$-- & 50000 & $\phantom{00}$54 & $\phantom{00}$721 & $\phantom{0}$721 & $\phantom{0}$3041 & 3041 \\
	HB \cite{kaskman2019homebreweddb} & $\ast$ & 33 & $\phantom{00000}$-- & 50000 & 4420 & 13000 & $\phantom{0}$300 & 67542 & 1630 \\
	YCB-V \cite{xiang2017posecnn} & $\ast$ & 21 & 113198 & 50000 & $\phantom{000}$-- & 20738 & $\phantom{0}$900 & 98547 & 4123 \\
	RU-APC \cite{rennie2016dataset} & & 14 & $\phantom{00000}$-- & $\phantom{0000}$-- & $\phantom{000}$-- & $\phantom{0}$5964 & 1380 & $\phantom{0}$5964 & 1380\\
	IC-BIN \cite{doumanoglou2016recovering} & $\ast$ & $\phantom{0}$2 & $\phantom{00000}$-- & 50000 & $\phantom{000}$-- & $\phantom{00}$177 & $\phantom{0}$150 & $\phantom{0}$2176 & 1786 \\
	IC-MI \cite{tejani2014latent} & & $\phantom{0}$6 & $\phantom{00000}$-- & $\phantom{0000}$-- & $\phantom{000}$-- & $\phantom{0}$2067 & $\phantom{0}$300 & $\phantom{0}$5318 & $\phantom{0}$800 \\
	TUD-L \cite{hodan2018bop} & $\ast$ & $\phantom{0}$3 & $\phantom{0}$38288 & 50000 & $\phantom{000}$-- & 23914 & $\phantom{0}$600 & 23914 & $\phantom{0}$600 \\
	TYO-L \cite{hodan2018bop} & & 21 & $\phantom{00000}$-- & $\phantom{0000}$-- & $\phantom{000}$-- & $\phantom{0}$1670 & 1670 & $\phantom{0}$1670 & 1670 \\
	\bottomrule
\end{tabularx}
\captionof{table}{\label{tab:dataset_params} \textbf{Parameters of the BOP datasets.} Most datasets include also training images obtained by OpenGL rendering of the 3D object models on a black background (not shown in the table).
Extra PBR training images can be rendered by BlenderProc4BOP~\cite{denninger2019blenderproc,denninger2020blenderproc}.
If a dataset includes both validation and test images, the ground-truth annotations are public only for the validation images. All test images are real. Column ``Test inst./All'' shows the number of annotated object instances for which at least $10\%$ of the projected surface area is visible in the test images. Columns ``Used'' show the number of test images and object instances used in the BOP Challenge 2019 \& 2020.
\vspace{7.8ex}
}
\endgroup

\begingroup
\small

\renewcommand{\arraystretch}{0.9}

\begin{tabular}{ @{}c@{ } @{}c@{ } @{}c@{ } @{}c@{ } }
LM~\cite{hinterstoisser2012accv} & LM-O*~\cite{brachmann2014learning} & T-LESS*~\cite{hodan2017tless} & ITODD*~\cite{drost2017introducing} \vspace{0.5ex} \\
\includegraphics[width=0.243\columnwidth]{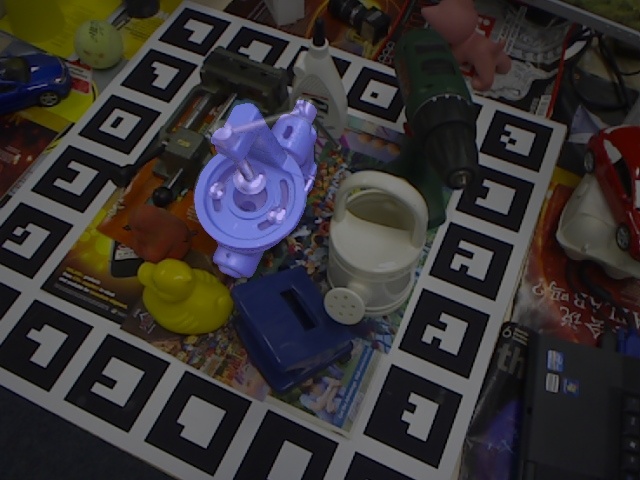} &
\includegraphics[width=0.243\columnwidth]{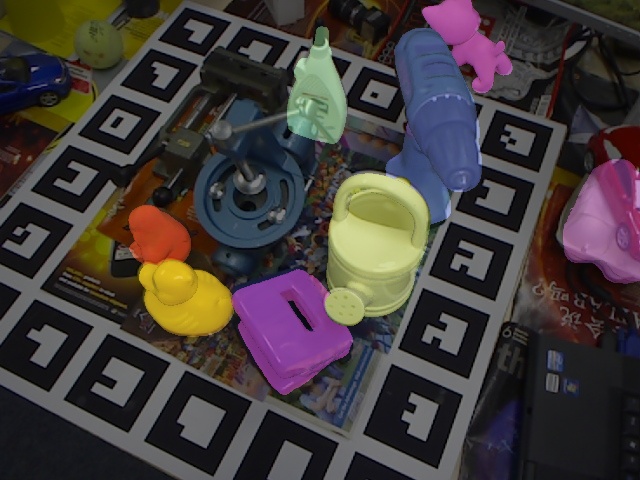} &
\includegraphics[width=0.243\columnwidth]{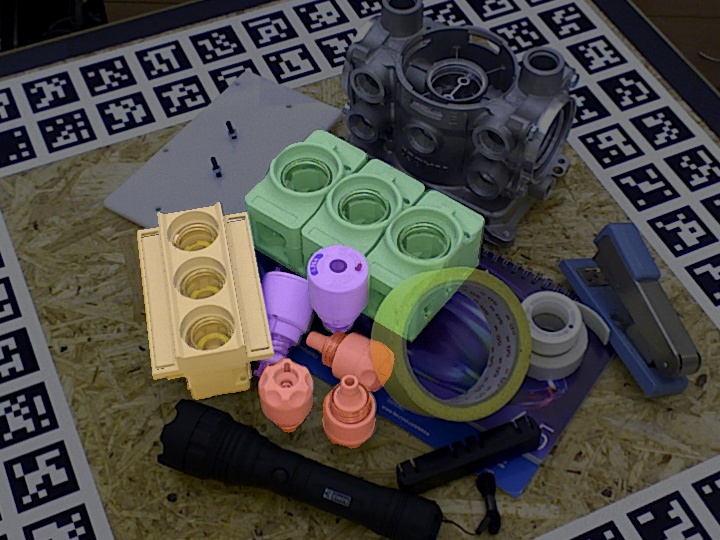} &
\includegraphics[width=0.243\columnwidth]{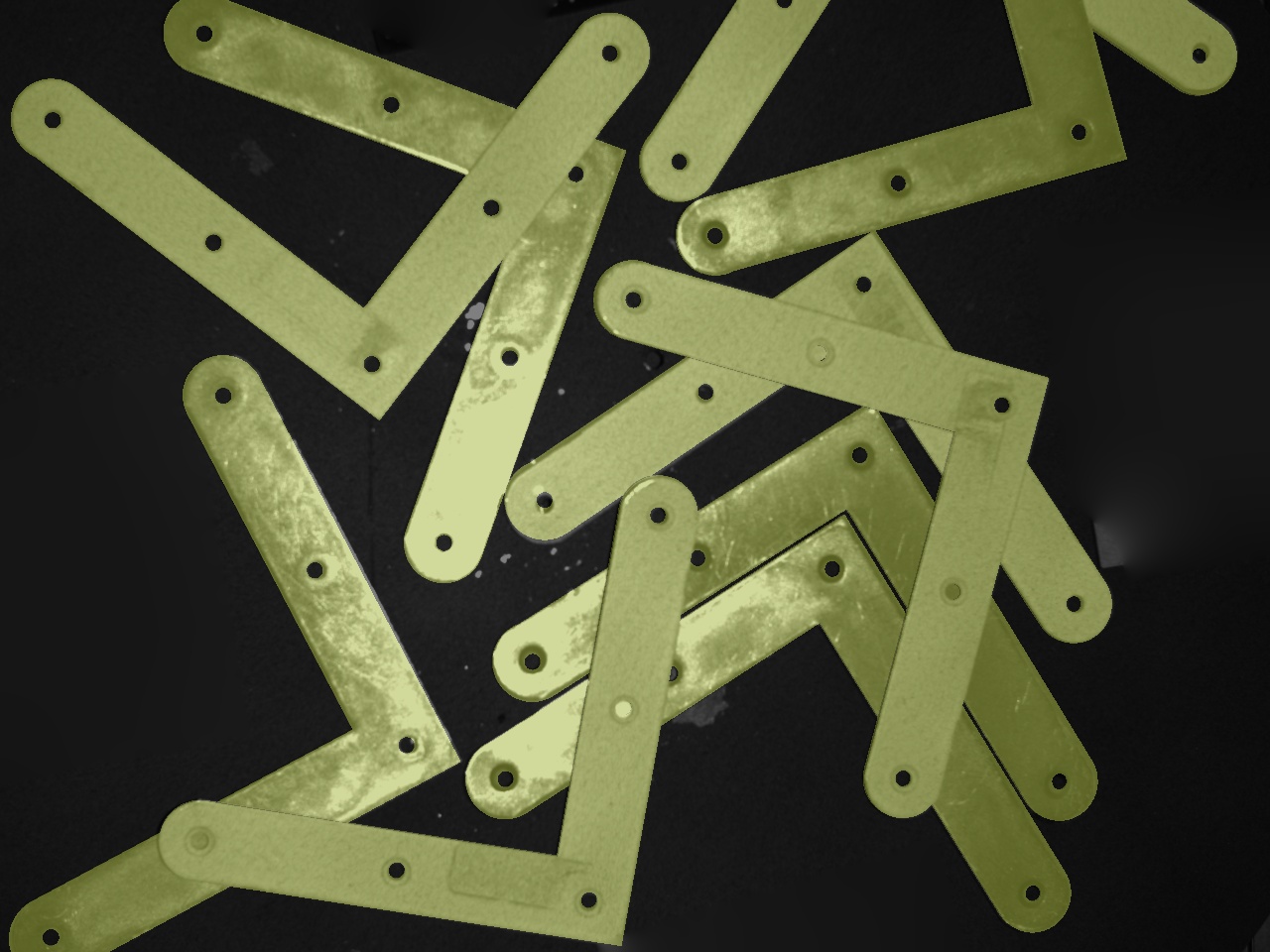} \vspace{0.1ex} \\
\end{tabular}

\begin{tabular}{ @{}c@{ } @{}c@{ } @{}c@{ } @{}c@{ } }
HB*~\cite{kaskman2019homebreweddb} & YCB-V*~\cite{xiang2017posecnn} & RU-APC~\cite{rennie2016dataset} & IC-BIN*~\cite{doumanoglou2016recovering} \vspace{0.5ex} \\
\includegraphics[width=0.243\columnwidth]{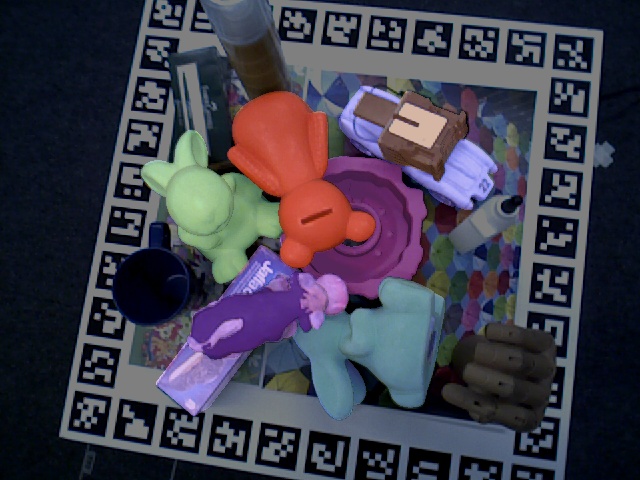} &
\includegraphics[width=0.243\columnwidth]{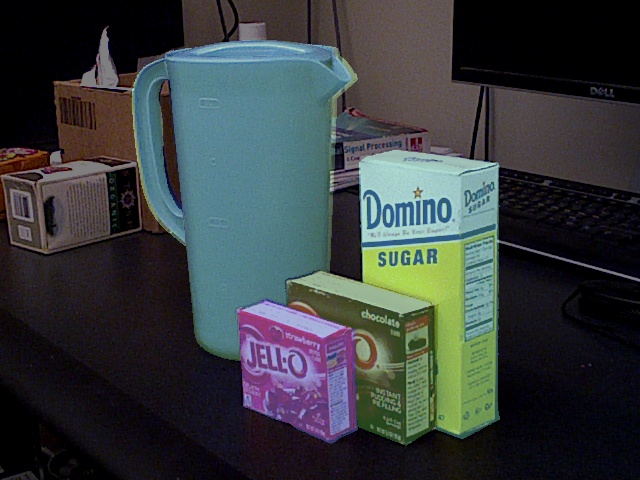} &
\includegraphics[width=0.243\columnwidth]{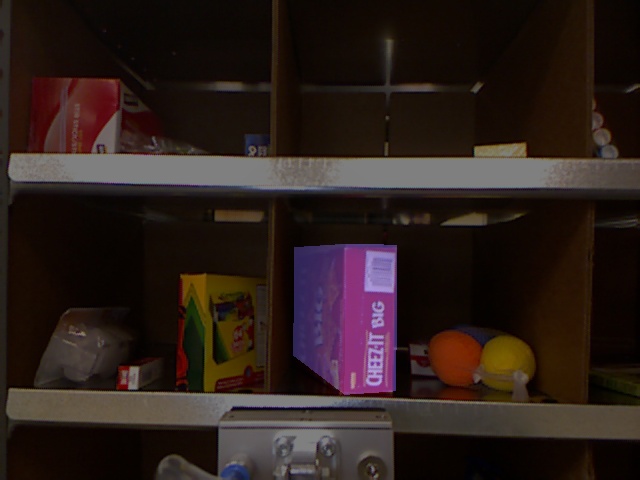} &
\includegraphics[width=0.243\columnwidth]{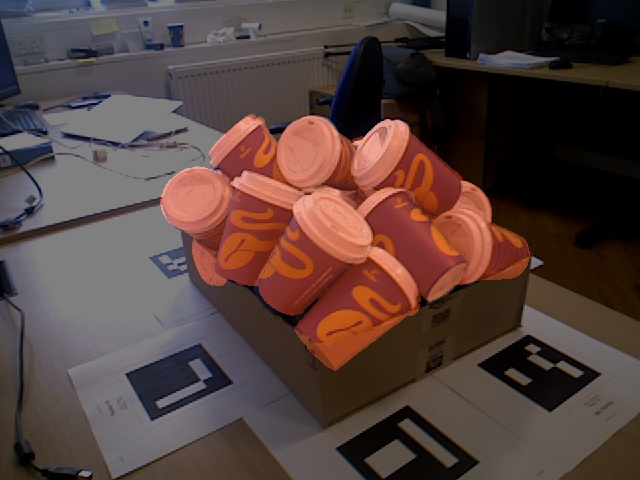} \vspace{0.1ex} \\
\end{tabular}

\begin{tabular}{ @{}c@{ } @{}c@{ } @{}c@{ } }
IC-MI~\cite{tejani2014latent} & TUD-L*~\cite{hodan2018bop} & TYO-L~\cite{hodan2018bop} \vspace{0.5ex} \\
\includegraphics[width=0.243\columnwidth]{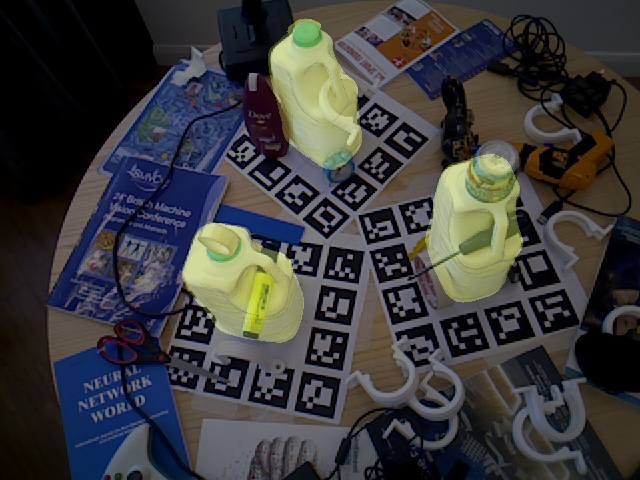} &
\includegraphics[width=0.243\columnwidth]{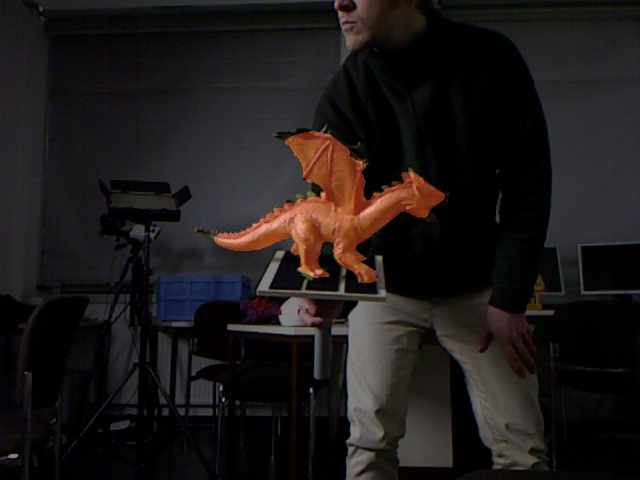} &
\includegraphics[width=0.243\columnwidth]{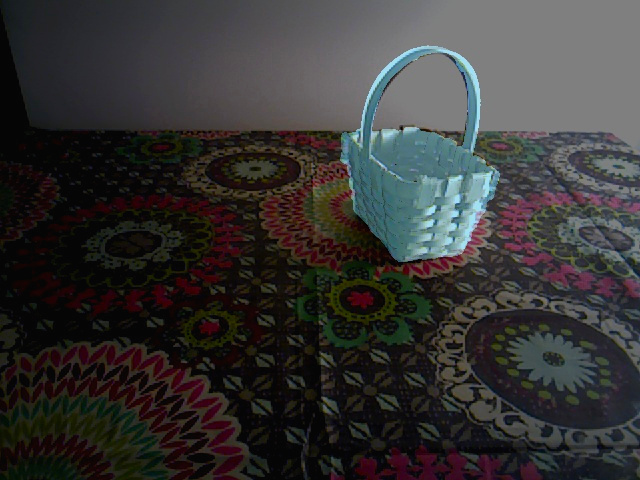} \\
\end{tabular}

\endgroup

\caption{\label{fig:core_datasets}
\textbf{An overview of the BOP datasets.} The core datasets are marked with a star. Shown are RGB channels of sample test images which were darkened and overlaid with colored 3D object models in the ground-truth 6D poses.}

\end{center}
\end{figure}

%% file: tab_results.tex
\setlength{\tabcolsep}{2.5pt}
\begin{table}[t!]

\begin{center}
\scriptsize
\begin{tabularx}{\linewidth}{r l Y Y Y Y Y Y Y Y Y}
\toprule
\# &
Method &
\rotatebox[origin=c]{90}{Avg.} &
\rotatebox[origin=c]{90}{LM-O} &
\rotatebox[origin=c]{90}{T-LESS} &
\rotatebox[origin=c]{90}{TUD-L} &
\rotatebox[origin=c]{90}{IC-BIN} &
\rotatebox[origin=c]{90}{ITODD} &
\rotatebox[origin=c]{90}{HB} &
\rotatebox[origin=c]{90}{YCB-V} &
\rotatebox[origin=c]{90}{Time} \\
\midrule    
1 & CosyPose-ECCV20-Synt+Real-ICP~\cite{labbe2020cosypose} & \cellcolor{avgcol!69.8}69.8 & \cellcolor{arcol!71.4}71.4 & \cellcolor{arcol!70.1}70.1 & \cellcolor{arcol!93.9}93.9 & \cellcolor{arcol!64.7}64.7 & \cellcolor{arcol!31.3}31.3 & \cellcolor{arcol!71.2}71.2 & \cellcolor{arcol!86.1}86.1 & \cellcolor{timecol!13.74}13.74 \\
2 & K\"onig-Hybrid-DL-PointPairs~\cite{koenig2020hybrid} & \cellcolor{avgcol!63.9}63.9 & \cellcolor{arcol!63.1}63.1 & \cellcolor{arcol!65.5}65.5 & \cellcolor{arcol!92.0}92.0 & \cellcolor{arcol!43.0}43.0 & \cellcolor{arcol!48.3}48.3 & \cellcolor{arcol!65.1}65.1 & \cellcolor{arcol!70.1}70.1 & $\phantom{0}$\cellcolor{timecol!0.63}0.63 \\
3 & CosyPose-ECCV20-Synt+Real~\cite{labbe2020cosypose} & \cellcolor{avgcol!63.7}63.7 & \cellcolor{arcol!63.3}63.3 & \cellcolor{arcol!72.8}72.8 & \cellcolor{arcol!82.3}82.3 & \cellcolor{arcol!58.3}58.3 & \cellcolor{arcol!21.6}21.6 & \cellcolor{arcol!65.6}65.6 & \cellcolor{arcol!82.1}82.1 & $\phantom{0}$\cellcolor{timecol!0.45}0.45 \\
4 & Pix2Pose-BOP20\_w/ICP-ICCV19~\cite{park2019pix2pose} & \cellcolor{avgcol!59.1}59.1 & \cellcolor{arcol!58.8}58.8 & \cellcolor{arcol!51.2}51.2 & \cellcolor{arcol!82.0}82.0 & \cellcolor{arcol!39.0}39.0 & \cellcolor{arcol!35.1}35.1 & \cellcolor{arcol!69.5}69.5 & \cellcolor{arcol!78.0}78.0 & $\phantom{0}$\cellcolor{timecol!4.84}4.84 \\
5 & CosyPose-ECCV20-PBR~\cite{labbe2020cosypose} & \cellcolor{avgcol!57.0}57.0 & \cellcolor{arcol!63.3}63.3 & \cellcolor{arcol!64.0}64.0 & \cellcolor{arcol!68.5}68.5 & \cellcolor{arcol!58.3}58.3 & \cellcolor{arcol!21.6}21.6 & \cellcolor{arcol!65.6}65.6 & \cellcolor{arcol!57.4}57.4 & $\phantom{0}$\cellcolor{timecol!0.47}0.47 \\
6 & Vidal-Sensors18~\cite{vidal2018method} & \cellcolor{avgcol!56.9}56.9 & \cellcolor{arcol!58.2}58.2 & \cellcolor{arcol!53.8}53.8 & \cellcolor{arcol!87.6}87.6 & \cellcolor{arcol!39.3}39.3 & \cellcolor{arcol!43.5}43.5 & \cellcolor{arcol!70.6}70.6 & \cellcolor{arcol!45.0}45.0 & $\phantom{0}$\cellcolor{timecol!3.22}3.22 \\
7 & CDPNv2\_BOP20-RGB-ICP~\cite{li2019cdpn} & \cellcolor{avgcol!56.8}56.8 & \cellcolor{arcol!63.0}63.0 & \cellcolor{arcol!46.4}46.4 & \cellcolor{arcol!91.3}91.3 & \cellcolor{arcol!45.0}45.0 & \cellcolor{arcol!18.6}18.6 & \cellcolor{arcol!71.2}71.2 & \cellcolor{arcol!61.9}61.9 & $\phantom{0}$\cellcolor{timecol!1.46}1.46 \\
8 & Drost-CVPR10-Edges~\cite{drost2010model} & \cellcolor{avgcol!55.0}55.0 & \cellcolor{arcol!51.5}51.5 & \cellcolor{arcol!50.0}50.0 & \cellcolor{arcol!85.1}85.1 & \cellcolor{arcol!36.8}36.8 & \cellcolor{arcol!57.0}57.0 & \cellcolor{arcol!67.1}67.1 & \cellcolor{arcol!37.5}37.5 & \cellcolor{timecol!87.57}87.57 \\
9 & CDPNv2\_BOP20-PBR-ICP~\cite{li2019cdpn} & \cellcolor{avgcol!53.4}53.4 & \cellcolor{arcol!63.0}63.0 & \cellcolor{arcol!43.5}43.5 & \cellcolor{arcol!79.1}79.1 & \cellcolor{arcol!45.0}45.0 & \cellcolor{arcol!18.6}18.6 & \cellcolor{arcol!71.2}71.2 & \cellcolor{arcol!53.2}53.2 & $\phantom{0}$\cellcolor{timecol!1.49}1.49 \\
10 & CDPNv2\_BOP20-RGB~\cite{li2019cdpn} & \cellcolor{avgcol!52.9}52.9 & \cellcolor{arcol!62.4}62.4 & \cellcolor{arcol!47.8}47.8 & \cellcolor{arcol!77.2}77.2 & \cellcolor{arcol!47.3}47.3 & \cellcolor{arcol!10.2}10.2 & \cellcolor{arcol!72.2}72.2 & \cellcolor{arcol!53.2}53.2 & $\phantom{0}$\cellcolor{timecol!0.94}0.94 \\
11 & Drost-CVPR10-3D-Edges~\cite{drost2010model} & \cellcolor{avgcol!50.0}50.0 & \cellcolor{arcol!46.9}46.9 & \cellcolor{arcol!40.4}40.4 & \cellcolor{arcol!85.2}85.2 & \cellcolor{arcol!37.3}37.3 & \cellcolor{arcol!46.2}46.2 & \cellcolor{arcol!62.3}62.3 & \cellcolor{arcol!31.6}31.6 & \cellcolor{timecol!80.06}80.06 \\
12 & Drost-CVPR10-3D-Only~\cite{drost2010model} & \cellcolor{avgcol!48.7}48.7 & \cellcolor{arcol!52.7}52.7 & \cellcolor{arcol!44.4}44.4 & \cellcolor{arcol!77.5}77.5 & \cellcolor{arcol!38.8}38.8 & \cellcolor{arcol!31.6}31.6 & \cellcolor{arcol!61.5}61.5 & \cellcolor{arcol!34.4}34.4 & $\phantom{0}$\cellcolor{timecol!7.70}7.70 \\
13 & CDPN\_BOP19-RGB~\cite{li2019cdpn} & \cellcolor{avgcol!47.9}47.9 & \cellcolor{arcol!56.9}56.9 & \cellcolor{arcol!49.0}49.0 & \cellcolor{arcol!76.9}76.9 & \cellcolor{arcol!32.7}32.7 & $\phantom{0}$\cellcolor{arcol!6.7}6.7 & \cellcolor{arcol!67.2}67.2 & \cellcolor{arcol!45.7}45.7 & $\phantom{0}$\cellcolor{timecol!0.48}0.48 \\
14 & CDPNv2\_BOP20-PBR~\cite{li2019cdpn} & \cellcolor{avgcol!47.2}47.2 & \cellcolor{arcol!62.4}62.4 & \cellcolor{arcol!40.7}40.7 & \cellcolor{arcol!58.8}58.8 & \cellcolor{arcol!47.3}47.3 & \cellcolor{arcol!10.2}10.2 & \cellcolor{arcol!72.2}72.2 & \cellcolor{arcol!39.0}39.0 & $\phantom{0}$\cellcolor{timecol!0.98}0.98 \\
15 & leaping from 2D to 6D~\cite{liu2010leaping} & \cellcolor{avgcol!47.1}47.1 & \cellcolor{arcol!52.5}52.5 & \cellcolor{arcol!40.3}40.3 & \cellcolor{arcol!75.1}75.1 & \cellcolor{arcol!34.2}34.2 & $\phantom{0}$\cellcolor{arcol!7.7}7.7 & \cellcolor{arcol!65.8}65.8 & \cellcolor{arcol!54.3}54.3 & $\phantom{0}$\cellcolor{timecol!0.42}0.42 \\
16 & EPOS-BOP20-PBR~\cite{hodan2020epos} & \cellcolor{avgcol!45.7}45.7 & \cellcolor{arcol!54.7}54.7 & \cellcolor{arcol!46.7}46.7 & \cellcolor{arcol!55.8}55.8 & \cellcolor{arcol!36.3}36.3 & \cellcolor{arcol!18.6}18.6 & \cellcolor{arcol!58.0}58.0 & \cellcolor{arcol!49.9}49.9 & $\phantom{0}$\cellcolor{timecol!1.87}1.87 \\
17 & Drost-CVPR10-3D-Only-Faster~\cite{drost2010model} & \cellcolor{avgcol!45.4}45.4 & \cellcolor{arcol!49.2}49.2 & \cellcolor{arcol!40.5}40.5 & \cellcolor{arcol!69.6}69.6 & \cellcolor{arcol!37.7}37.7 & \cellcolor{arcol!27.4}27.4 & \cellcolor{arcol!60.3}60.3 & \cellcolor{arcol!33.0}33.0 & $\phantom{0}$\cellcolor{timecol!1.38}1.38 \\
18 & F{\'e}lix\&Neves-ICRA17-IET19~\cite{rodrigues2019deep,raposo2017using} & \cellcolor{avgcol!41.2}41.2 & \cellcolor{arcol!39.4}39.4 & \cellcolor{arcol!21.2}21.2 & \cellcolor{arcol!85.1}85.1 & \cellcolor{arcol!32.3}32.3 & $\phantom{0}$\cellcolor{arcol!6.9}6.9 & \cellcolor{arcol!52.9}52.9 & \cellcolor{arcol!51.0}51.0 & \cellcolor{timecol!55.78}55.78 \\
19 & Sundermeyer-IJCV19+ICP~\cite{sundermeyer2019augmented} & \cellcolor{avgcol!39.8}39.8 & \cellcolor{arcol!23.7}23.7 & \cellcolor{arcol!48.7}48.7 & \cellcolor{arcol!61.4}61.4 & \cellcolor{arcol!28.1}28.1 & \cellcolor{arcol!15.8}15.8 & \cellcolor{arcol!50.6}50.6 & \cellcolor{arcol!50.5}50.5 & $\phantom{0}$\cellcolor{timecol!0.86}0.86 \\
20 & Zhigang-CDPN-ICCV19~\cite{li2019cdpn} & \cellcolor{avgcol!35.3}35.3 & \cellcolor{arcol!37.4}37.4 & \cellcolor{arcol!12.4}12.4 & \cellcolor{arcol!75.7}75.7 & \cellcolor{arcol!25.7}25.7 & $\phantom{0}$\cellcolor{arcol!7.0}7.0 & \cellcolor{arcol!47.0}47.0 & \cellcolor{arcol!42.2}42.2 & $\phantom{0}$\cellcolor{timecol!0.51}0.51 \\
21 & PointVoteNet2~\cite{hagelskjaer2019pointposenet} & \cellcolor{avgcol!35.1}35.1 & \cellcolor{arcol!65.3}65.3 & $\phantom{0}$\cellcolor{arcol!0.4}0.4 & \cellcolor{arcol!67.3}67.3 & \cellcolor{arcol!26.4}26.4 & $\phantom{0}$\cellcolor{arcol!0.1}0.1 & \cellcolor{arcol!55.6}55.6 & \cellcolor{arcol!30.8}30.8 & \cellcolor{timecol!0}- \\
22 & Pix2Pose-BOP20-ICCV19~\cite{park2019pix2pose} & \cellcolor{avgcol!34.2}34.2 & \cellcolor{arcol!36.3}36.3 & \cellcolor{arcol!34.4}34.4 & \cellcolor{arcol!42.0}42.0 & \cellcolor{arcol!22.6}22.6 & \cellcolor{arcol!13.4}13.4 & \cellcolor{arcol!44.6}44.6 & \cellcolor{arcol!45.7}45.7 & $\phantom{0}$\cellcolor{timecol!1.22}1.22 \\
23 & Sundermeyer-IJCV19~\cite{sundermeyer2019augmented} & \cellcolor{avgcol!27.0}27.0 & \cellcolor{arcol!14.6}14.6 & \cellcolor{arcol!30.4}30.4 & \cellcolor{arcol!40.1}40.1 & \cellcolor{arcol!21.7}21.7 & \cellcolor{arcol!10.1}10.1 & \cellcolor{arcol!34.6}34.6 & \cellcolor{arcol!37.7}37.7 & $\phantom{0}$\cellcolor{timecol!0.19}0.19 \\
24 & SingleMultiPathEncoder-CVPR20~\cite{sundermeyer2020multi} & \cellcolor{avgcol!24.1}24.1 & \cellcolor{arcol!21.7}21.7 & \cellcolor{arcol!31.0}31.0 & \cellcolor{arcol!33.4}33.4 & \cellcolor{arcol!17.5}17.5 & $\phantom{0}$\cellcolor{arcol!6.7}6.7 & \cellcolor{arcol!29.3}29.3 & \cellcolor{arcol!28.9}28.9 & $\phantom{0}$\cellcolor{timecol!0.19}0.19 \\
25 & Pix2Pose-BOP19-ICCV19~\cite{park2019pix2pose} & \cellcolor{avgcol!20.5}20.5 & $\phantom{0}$\cellcolor{arcol!7.7}7.7 & \cellcolor{arcol!27.5}27.5 & \cellcolor{arcol!34.9}34.9 & \cellcolor{arcol!21.5}21.5 & $\phantom{0}$\cellcolor{arcol!3.2}3.2 & \cellcolor{arcol!20.0}20.0 & \cellcolor{arcol!29.0}29.0 & $\phantom{0}$\cellcolor{timecol!0.79}0.79 \\
26 & DPOD (synthetic)~\cite{zakharov2019dpod} & \cellcolor{avgcol!16.1}16.1 & \cellcolor{arcol!16.9}16.9 & $\phantom{0}$\cellcolor{arcol!8.1}8.1 & \cellcolor{arcol!24.2}24.2 & \cellcolor{arcol!13.0}13.0 & $\phantom{0}$\cellcolor{arcol!0.0}0.0 & \cellcolor{arcol!28.6}28.6 & \cellcolor{arcol!22.2}22.2 & $\phantom{0}$\cellcolor{timecol!0.23}0.23 \\

\\
\end{tabularx}

\begin{tabularx}{\linewidth}{r l l l l l l L l}
\toprule
\# &
Method &
Year &
PPF &
DNN &
Train &
...type &
Test &
Refine \\
\midrule    
1 & CosyPose-ECCV20-Synt+Real-ICP~\cite{labbe2020cosypose} & \cellcolor{ccol!100}2020 & - & \cellcolor{ccol!100}3/set & \cellcolor{ccol!100}rgb & pbr{\tiny +}real & rgb-d & \cellcolor{ccol!100}rgb{\tiny +}icp \\
2 & K\"onig-Hybrid-DL-PointPairs~\cite{koenig2020hybrid} & \cellcolor{ccol!100}2020 & \cellcolor{ccol!100}yes & \cellcolor{ccol!100}1/set & \cellcolor{ccol!100}rgb & syn{\tiny +}real & rgb-d & \cellcolor{ccol!100}icp \\
3 & CosyPose-ECCV20-Synt+Real~\cite{labbe2020cosypose} & \cellcolor{ccol!100}2020 & - & \cellcolor{ccol!100}3/set & \cellcolor{ccol!100}rgb & pbr{\tiny +}real & \cellcolor{ccol!100}rgb & \cellcolor{ccol!100}rgb \\
4 & Pix2Pose-BOP20\_w/ICP-ICCV19~\cite{park2019pix2pose} & \cellcolor{ccol!100}2020 & - & 1/obj & \cellcolor{ccol!100}rgb & pbr{\tiny +}real & rgb-d & \cellcolor{ccol!100}icp \\
5 & CosyPose-ECCV20-PBR~\cite{labbe2020cosypose} & \cellcolor{ccol!100}2020 & - & \cellcolor{ccol!100}3/set & \cellcolor{ccol!100}rgb & \cellcolor{ccol!100}pbr & \cellcolor{ccol!100}rgb & \cellcolor{ccol!100}rgb \\
6 & Vidal-Sensors18~\cite{vidal2018method} & 2019 & \cellcolor{ccol!100}yes & - & - & - & d & \cellcolor{ccol!100}icp \\
7 & CDPNv2\_BOP20-RGB-ICP~\cite{li2019cdpn} & \cellcolor{ccol!100}2020 & - & 1/obj & \cellcolor{ccol!100}rgb & pbr{\tiny +}real & rgb-d & \cellcolor{ccol!100}icp \\
8 & Drost-CVPR10-Edges~\cite{drost2010model} & 2019 & \cellcolor{ccol!100}yes & - & - & - & rgb-d & \cellcolor{ccol!100}icp \\
9 & CDPNv2\_BOP20-PBR-ICP~\cite{li2019cdpn} & \cellcolor{ccol!100}2020 & - & 1/obj & \cellcolor{ccol!100}rgb & \cellcolor{ccol!100}pbr & rgb-d & \cellcolor{ccol!100}icp \\
10 & CDPNv2\_BOP20-RGB~\cite{li2019cdpn} & \cellcolor{ccol!100}2020 & - & 1/obj & \cellcolor{ccol!100}rgb & pbr{\tiny +}real & \cellcolor{ccol!100}rgb & - \\
11 & Drost-CVPR10-3D-Edges~\cite{drost2010model} & 2019 & \cellcolor{ccol!100}yes & - & - & - & d & \cellcolor{ccol!100}icp \\
12 & Drost-CVPR10-3D-Only~\cite{drost2010model} & 2019 & \cellcolor{ccol!100}yes & - & - & - & d & \cellcolor{ccol!100}icp \\
13 & CDPN\_BOP19-RGB~\cite{li2019cdpn} & \cellcolor{ccol!100}2020 & - & 1/obj & \cellcolor{ccol!100}rgb & pbr{\tiny +}real & \cellcolor{ccol!100}rgb & - \\
14 & CDPNv2\_BOP20-PBR~\cite{li2019cdpn} & \cellcolor{ccol!100}2020 & - & 1/obj & \cellcolor{ccol!100}rgb & \cellcolor{ccol!100}pbr & \cellcolor{ccol!100}rgb & - \\
15 & leaping from 2D to 6D~\cite{liu2010leaping} & \cellcolor{ccol!100}2020 & - & 1/obj & \cellcolor{ccol!100}rgb & pbr{\tiny +}real & \cellcolor{ccol!100}rgb & - \\
16 & EPOS-BOP20-PBR~\cite{hodan2020epos} & \cellcolor{ccol!100}2020 & - & \cellcolor{ccol!100}1/set & \cellcolor{ccol!100}rgb & \cellcolor{ccol!100}pbr & \cellcolor{ccol!100}rgb & - \\
17 & Drost-CVPR10-3D-Only-Faster~\cite{drost2010model} & 2019 & \cellcolor{ccol!100}yes & - & - & - & d & \cellcolor{ccol!100}icp \\
18 & F{\'e}lix\&Neves-ICRA17-IET19~\cite{rodrigues2019deep,raposo2017using} & 2019 & \cellcolor{ccol!100}yes & \cellcolor{ccol!100}1/set & rgb-d & syn{\tiny +}real & rgb-d & \cellcolor{ccol!100}icp \\
19 & Sundermeyer-IJCV19+ICP~\cite{sundermeyer2019augmented} & 2019 & - & 1/obj & \cellcolor{ccol!100}rgb & syn{\tiny +}real & rgb-d & \cellcolor{ccol!100}icp \\
20 & Zhigang-CDPN-ICCV19~\cite{li2019cdpn} & 2019 & - & 1/obj & \cellcolor{ccol!100}rgb & syn{\tiny +}real & \cellcolor{ccol!100}rgb & - \\
21 & PointVoteNet2~\cite{hagelskjaer2019pointposenet} & \cellcolor{ccol!100}2020 & - & 1/obj & rgb-d & \cellcolor{ccol!100}pbr & rgb-d & \cellcolor{ccol!100}icp \\
22 & Pix2Pose-BOP20-ICCV19~\cite{park2019pix2pose} & \cellcolor{ccol!100}2020 & - & 1/obj & \cellcolor{ccol!100}rgb & pbr{\tiny +}real & \cellcolor{ccol!100}rgb & - \\
23 & Sundermeyer-IJCV19~\cite{sundermeyer2019augmented} & 2019 & - & 1/obj & \cellcolor{ccol!100}rgb & syn{\tiny +}real & \cellcolor{ccol!100}rgb & - \\
24 & SingleMultiPathEncoder-CVPR20~\cite{sundermeyer2020multi} & \cellcolor{ccol!100}2020 & - & \cellcolor{ccol!100}1/all & \cellcolor{ccol!100}rgb & syn{\tiny +}real & \cellcolor{ccol!100}rgb & - \\
25 & Pix2Pose-BOP19-ICCV19~\cite{park2019pix2pose} & 2019 & - & 1/obj & \cellcolor{ccol!100}rgb & syn{\tiny +}real & \cellcolor{ccol!100}rgb & - \\
26 & DPOD (synthetic)~\cite{zakharov2019dpod} & 2019 & - & \cellcolor{ccol!100}1/scene & \cellcolor{ccol!100}rgb & syn & \cellcolor{ccol!100}rgb & - \\

\bottomrule
\end{tabularx}
\end{center}

\caption{\label{tab:results} \textbf{Results of the BOP Challenge 2019 and 2020.} The methods are ranked by the $\text{AR}_{\text{Core}}$ score (the third column of the upper table) which is the average of the per-dataset $\text{AR}_D$ scores (the following seven columns). The scores are defined in Sec.~\ref{sec:accuracy_score}. The last column of the upper table shows the average image processing time~[s] averaged over the datasets. The lower table shows properties discussed in Sec.~\ref{sec:evaluation}.}

\end{table}

%% file: tab_cosypose_data.tex
\setlength{\tabcolsep}{5pt}
\begin{table}[t]
\centering

\begingroup
\footnotesize
\begin{center}
\begin{tabularx}{\textwidth}{ c c *{3}{Y} }
	\toprule
	Detection & Pose estim. & T-LESS & TUD-L & YCB-V \\
	\midrule
	PBR+Real & PBR+Real & 72.8 & 82.3 & 82.1 \\
	PBR & PBR & 64.0 & 68.5 & 57.4 \\
	PBR & Render\,\&\,paste v1 & 16.1 & 60.4 & 44.9 \\
	PBR & Render\,\&\,paste v2 & 60.0 & 58.9 & 58.5 \\
	Render\,\&\,paste v1 & Render\,\&\,paste v1 & 6.1 & 49.5 & 26.5 \\
	Render\,\&\,paste v2 & Render\,\&\,paste v2 & 45.3 & 42.4 & 25.7 \\
	\bottomrule
\end{tabularx}
\end{center}

\caption{\label{tab:cosy_data_res} \textbf{The effect of different training images}. The table shows the $\text{AR}_{\text{Core}}$ scores achieved by the CosyPose method~\cite{labbe2020cosypose} when different types of images were used for training its object detection (\ie Mask R-CNN~\cite{he2017mask}) and pose estimation stage.
The ``render\;\&\;paste v1'' images were obtained by OpenGL rendering of the 3D object models on random real photographs.
The ``render\;\&\;paste v2'' images were obtained similarly, but the CAD models of T-LESS objects were assigned a random surface texture instead of a random gray value, the background of most images was assigned a synthetic texture, and 1M instead of 50K images were generated.
Interestingly, the increased photorealism brought by the PBR images yields noticeable improvements despite the strong data augmentation applied by CosyPose to the training images.
\vspace{5.3ex}
}
\endgroup

\end{table}